\documentclass{article}
\usepackage{spconf,amsmath,graphicx,hyperref}

\usepackage{cite}
\usepackage{xurl}
\usepackage{hyperref}
\usepackage{graphicx}
\usepackage{textcomp}
\usepackage[dvipsnames]{xcolor}
\usepackage{url}
\usepackage{booktabs}
\usepackage{amsfonts}
\usepackage{nicefrac}
\usepackage{microtype}
\usepackage{subcaption}
\usepackage{array}
\usepackage{algorithm}
\usepackage{algorithmicx}
\usepackage{algpseudocode}
\usepackage{amsmath}
\usepackage{amssymb}
\usepackage{mathtools}
\usepackage{amsthm}
\usepackage{extpfeil}
\usepackage{multirow}
\usepackage{tikz}
\usepackage{makecell}
\usepackage{siunitx}

\usetikzlibrary{quantikz2}
\usetikzlibrary{calc, decorations.pathreplacing}
\usetikzlibrary{positioning, fit, backgrounds, arrows.meta}

\usepackage{thmtools}
\usepackage{mathrsfs}
\usepackage{tabularx}
\usepackage[most]{tcolorbox}
\usepackage{enumerate}
\usepackage{paralist}
\usepackage{enumitem}

\DeclarePairedDelimiterX{\norm}[1]{\lVert}{\rVert}{#1}

\theoremstyle{definition}

\theoremstyle{remark}

\title{Learning to Program Adaptive Non-Local Observables for Machine Learning}
\name{
  Yu-Ting Lee$^{1}$ \quad
  Samuel Yen-Chi Chen$^{2}$\thanks{The views expressed in this article are those of the authors and do not represent the views of Wells Fargo. This article is for informational purposes only. Nothing contained in this article should be construed as investment advice. Wells Fargo makes no express or implied warranties and expressly disclaims all legal, tax, and accounting implications related to this article.} \quad
  Huan-Hsin Tseng$^{3}$
}
\address{
  $^{1}$Graduate Institute of Communication Engineering, National Taiwan University, Taipei, Taiwan \\
  $^{2}$Wells Fargo, New York, NY, USA \\
  $^{3}$Brookhaven National Laboratory, AI \& ML Department, Upton, NY, USA \\
  r14942088@ntu.edu.tw, yen-chi.chen@wellsfargo.com, htseng@bnl.gov
  }

\begin{document}
\maketitle
\begin{abstract}
Quantum neural networks (QNNs) are typically built from variational quantum circuits (VQCs), which are limited by local measurements. Adaptive non-local observables (ANO) address this by jointly optimizing circuit parameters and multi-qubit measurements. However, existing ANO-based VQCs learn only a single static observable that remains invariant across all inputs. We propose QFWP-ANO, a novel architecture which employs a classical hypernetwork to dynamically program VQC parameters and/or non-local observables conditioned on each input. On multivariate time-series forecasting across four ETT datasets, QFWP-ANO achieves the lowest MSE in 16 of 20 settings and second-lowest in the remaining four, surpassing ANO-based and other strong baselines. On reinforcement learning tasks, QFWP-ANO consistently surpasses ANO-VQCs. Our results establish input-conditioned ANO as an effective approach for enhancing QNNs.
\end{abstract}
\section{Introduction}
Quantum machine learning (QML) and quantum neural networks (QNNs) have emerged as a promising paradigm that leverages the representational expressivity of quantum phenomena such as superposition, entanglement, and quantum interference~\cite{Cerezo2021, McClean_2016}. In particular, QNNs are increasingly being applied to a wide range of complex machine learning tasks, including reinforcement learning~\cite{NEURIPS2021_eec96a7f, 10889145, lee2026quantumhierarchicalreinforcementlearning, Chen:2020opi}, classification~\cite{farhi2018classificationquantumneuralnetworks, PerezSalinas2020datareuploading}, anomaly detection~\cite{11464824}, and time-series prediction~\cite{qaio25, 9747369, 10650743, 11461293, lee2026multivariatetimeseriesforecasting}.

However, QNNs are typically built from variational quantum circuits (VQCs) and depend on local measurements, which limit the network's ability to learn complex data distributions. To address this bottleneck, recent work proposes to jointly optimize circuit parameters  with trainable observables~\cite{11011001}. Specifically, the adaptive non-local observables (ANO) framework~\cite{11249836} leverages trainable multi-qubit Hermitian observables and shows strong potential across super-resolution~\cite{11463496}, reinforcement learning~\cite{11250031}, classification~\cite{11249836}, and time-series forecasting~\cite{lee2026multivariatetimeseriesforecasting}. Yet, a key limitation of existing ANO-based QNNs is their reliance on a static observable per task, rendering the measurement invariant to the input data stream.

To overcome this limitation, we propose QFWP-ANO, which renders non-local observables adaptive to the input. Building on quantum fast weight programmers (QFWP)~\cite{10650743}, our approach employs a classical hypernetwork to program the circuit parameters, the non-local observables, or both, yielding a family of input-conditioned models. Experiments on multivariate time-series forecasting (MTSF) and reinforcement learning (RL) demonstrate the superior performance of QFWP-ANO over standard ANO-VQCs. Our contributions:

\begin{itemize}
    \item We introduce QFWP-ANO, a novel QNN architecture that utilizes a classical hypernetwork to dynamically program VQCs along with non-local observables.
    \item On multivariate time-series forecasting across four ETT datasets, QFWP-ANO achieves the lowest MSE in 16 of 20 settings and second-lowest in the remaining four, outperforming standard ANO-VQCs in 17 of 20 cases.
    \item In standard RL environments, QFWP-ANO outperforms ANO-VQCs, while programming the non-local observables can accelerate learning.
\end{itemize}

\section{Methodology}

\subsection{Variational Quantum Circuits}
Variational quantum circuits (VQCs), also known as parameterized quantum circuits (PQCs), are trainable quantum models that process classical data in three stages. Initially, a classical input $x$ is mapped into an $n$-qubit system via a data encoding unitary circuit $U(x)$, yielding the encoded states $U(x)|0\rangle^{\otimes n}$, where $|0\rangle^{\otimes n}$ is the ground state. Subsequently, a parameterized unitary circuit $V(\theta)$ evolves this state into $V(\theta)U(x)|0\rangle^{\otimes n}$. This variational circuit $V(\theta)$ is typically structured with alternating layers of trainable single-qubit rotations and multi-qubit entangling gates. Lastly, a measurement layer extracts classical information by calculating the expectation values of a fixed Hermitian observable $H$. The computation of a VQC can be summarized as a quantum function $f_{\text{VQC}}(x;\theta)$:
\begin{equation}
f_{\text{VQC}}(x;\theta) = \langle 0|^{\otimes n} U^\dagger(x) V^\dagger(\theta) H V(\theta) U(x) |0\rangle^{\otimes n}.
\end{equation}

\subsection{Adaptive Non-Local Observables}

Motivated by the Heisenberg picture, where quantum evolution is characterized by dynamical observables, adaptive non-local observables (ANO)~\cite{11249836} replace the fixed observable of a standard VQC with a trainable Hermitian operator $H(\phi)$ parameterized by $\phi$. A $k$-local observable takes the form:
\begin{equation}\label{tag:k-local}
H(\phi) =
\begin{pmatrix}
c_{11} & a_{12} + i b_{12} & a_{13} + i b_{13} & \cdots & a_{1K} + i b_{1K} \\
* & c_{22} & a_{23} + i b_{23} & \cdots & a_{2K} + i b_{2K} \\
* & * & c_{33} & \cdots & a_{3K} + i b_{3K} \\
\vdots & \vdots & \vdots & \ddots & \vdots \\
* & * & * & \cdots & c_{KK}
\end{pmatrix}
\end{equation}
where $k \leq n$, $K = 2^k$, and the parameter set $\phi = (a_{ij}, b_{ij}, c_{ii})_{i,j=1}^K$ consists of $K^2$ real parameters.

Utilizing a trainable observable strictly expands a QNN's function class: a conventional VQC with a fixed local observable is a special case of ANO~\cite{11249836}. Moreover, a $k$-local observable $H(\phi)$ operates jointly across $k$ qubits, coupling features between distant qubits to facilitate an information mixture that common Pauli-Z measurements cannot capture.


\subsection{QFWP-ANO}

QFWP-ANO employs a classical neural network as a hypernetwork to program the rotation angles of the variational circuit and/or the ANO parameters, conditioned on each input. Specifically, this hypernetwork utilizes an encoder-decoder architecture. A shared MLP encoder first processes the input into a latent representation $\boldsymbol{z}$. Taking $\boldsymbol{z}$ as input, specific decoders then generate the necessary parameters: a linear layer for the rotation angles $\theta$, and a separate two-layer MLP for the ANO parameters $\phi$. In contrast to QFWP, which stores temporal memory by recurrently accumulating updates, QFWP-ANO maps each input independently, carrying no accumulated state.

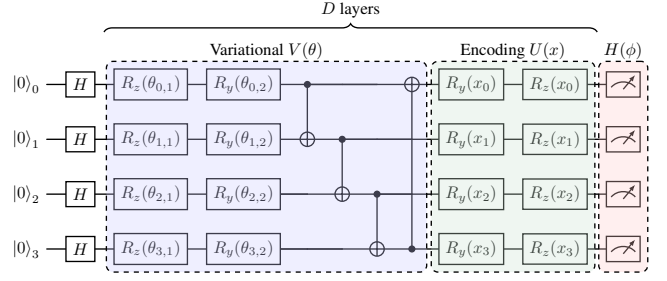
\begin{figure}[t!]
    \centering
\resizebox{\columnwidth}{!}{
\begin{quantikz}[
    column sep=0.4cm,
    execute at end picture={
        \draw[thick, decorate, decoration={brace, amplitude=5pt}] 
            ([yshift=2pt]Lgroup.north west) -- ([yshift=2pt]Lgroup.north east) 
            node[midway, above=5pt] {$D$ layers};
    }
]
    \lstick{$\ket{0}_0$} & \gate{H} & \gate{R_z(\theta_{0,1})} \gategroup[4,steps=6,style={dashed,rounded corners,inner sep=1pt,fill=blue!20,fill opacity=0.3}]{Variational $V(\theta)$} \gategroup[4,steps=8,style={draw=none, inner ysep=16pt, inner xsep=2pt, name=Lgroup}]{} & \gate{R_y(\theta_{0,2})} & \ctrl{1} & \qw & \qw & \targ{} & \gate{R_y(x_{0})} \gategroup[4,steps=2,style={dashed,rounded corners,inner sep=1pt,fill=ForestGreen!20,fill opacity=0.3}]{Encoding $U(x)$} & \gate{R_z(x_{0})} & \meter{} \gategroup[4,steps=1,style={dashed,rounded corners,inner sep=1pt,fill=red!20,fill opacity=0.3}]{$H(\phi)$} \\
    \lstick{$\ket{0}_1$} & \gate{H} & \gate{R_z(\theta_{1,1})} & \gate{R_y(\theta_{1,2})} & \targ{} & \ctrl{1} & \qw & \qw & \gate{R_y(x_{1})} & \gate{R_z(x_{1})} & \meter{} \\
    \lstick{$\ket{0}_2$} & \gate{H} & \gate{R_z(\theta_{2,1})} & \gate{R_y(\theta_{2,2})} & \qw & \targ{} & \ctrl{1} & \qw & \gate{R_y(x_{2})} & \gate{R_z(x_{2})} & \meter{} \\
    \lstick{$\ket{0}_3$} & \gate{H} & \gate{R_z(\theta_{3,1})} & \gate{R_y(\theta_{3,2})} & \qw & \qw & \targ{} & \ctrl{-3} & \gate{R_y(x_{3})} & \gate{R_z(x_{3})} & \meter{}
\end{quantikz}
}
\caption{\textbf{VQC architecture of QFWP-ANO.} Each layer consists parameterized $R_z$, $R_y$ rotations, circular CNOT gates, and encoding $U(x)$. Trainable $k$-local observables $H(\phi)$ are employed for measurement.}
\label{fig:vqc_architecture}
\end{figure}

We use a data re-uploading VQC structure~\cite{PerezSalinas2020datareuploading, PhysRevA.103.032430} combined with ANO. Initialized by a layer of Hadamard gates, each of the $D$ layers consists of 3 parts: parameterized $R_z$ and $R_y$ rotation gates, a circular topology of CNOT gates, and $R_y$ and $R_z$ encoding gates for data re-uploading (Fig.~\ref{fig:vqc_architecture}). Finally, a combinatorial measurement scheme with $k$-local observables is applied. We compute expected values across all $\binom{n}{k}$ combinations of $k$ qubits from the $n$ available; this generates one output value per combination to account for multi-qubit correlations.

\begin{table}[t!]
\centering
\caption{\textbf{Statistics of the datasets and environments.}}
\label{tab:stats}
\begin{subtable}{\columnwidth}
\centering
\caption{The four ETT datasets.}
\label{tab:stats_mtsf}
\begin{tabular}{lccc}
\toprule
Datasets & ETTh1 \& ETTh2 & ETTm1 \& ETTm2 \\
\midrule
Variates    & 7   & 7  \\
Timesteps   & 17,420 & 69,680 \\
Sample rate & 1 hour & 5 min \\
\bottomrule
\end{tabular}
\end{subtable}

\vspace{1em}

\begin{subtable}{\columnwidth}
\centering
\caption{RL environments.}
\label{tab:stats_rl}
\begin{tabular}{lcc}
\toprule
Environments & Acrobot & SimpleCrossingS9N1 \\
\midrule
State Space & Continuous & Discrete \\
Action Space & Discrete & Discrete \\
State dim    & 6 & 147 \\
Actions      & 3 & 7 \\
Reward       & $-1$/step & $+1$ on goal, else $0$\\
\bottomrule
\end{tabular}
\end{subtable}
\end{table}

\section{Experimental Settings}

We evaluate three QFWP-ANO variants in which the hypernetwork programs only the
rotation angles (QFWP-ANO-R), only the ANO parameters (QFWP-ANO-O), or both (QFWP-ANO-RO). We test
QFWP-ANO on two common QML tasks: multivariate time-series forecasting (MTSF) and RL.

\subsection{Multivariate Time Series Forecasting}

For a multivariate series with $C$ variates (channels), let $L$ be the lookback
window and $H$ the forecasting horizon. Given historical data
$\mathbf{X}_t \in \mathbb{R}^{C\times L}$, MTSF aims to predict future values $\mathbf{\widehat{Y}}_t \in \mathbb{R}^{C\times H}$. The  ground truth is denoted $\mathbf{Y}_t \in \mathbb{R}^{C\times H}$.

Following Lee et al.~\cite{lee2026multivariatetimeseriesforecasting}, we apply instance
normalization~\cite{kim2022reversible} to each channel of $X_t$ against
distribution shift:
\begin{equation}
\mathbf{X}'_t = (\mathbf{X}_t - \boldsymbol{\mu})\odot(\boldsymbol{\sigma}^2+\boldsymbol{\epsilon})^{-1/2},
\end{equation}
where $\boldsymbol{\mu},\boldsymbol{\sigma}^2\in\mathbb{R}^C$ are the per-channel mean and variance
over the lookback window and $\boldsymbol{\epsilon}$ ensures numerical stability. The
normalized input is flattened and projected to a latent $\boldsymbol{h}_t$, transformed
by QFWP-ANO, and mapped to forecasts via a linear layer and
de-normalization:
\begin{equation}
\hat{Y}_t = \big(W_{\text{out}} \boldsymbol{f}_{\text{VQC}}(\boldsymbol{h}_t;\theta,\phi)
+ \boldsymbol{b}_{\text{out}}\big)\odot\sqrt{\boldsymbol{\sigma}^2+\boldsymbol{\epsilon}}+\boldsymbol{\mu},
\end{equation}
where $\boldsymbol{f}_{\text{VQC}}(\boldsymbol{h}_t;\theta,\phi)\in\mathbb{R}^{\binom{n}{k}}$.

We conduct experiments on four real-world datasets from the Electricity Transformer Temperature (ETT) benchmark~\cite{Zhou_Zhang_Peng_Zhang_Li_Xiong_Zhang_2021} (Table~\ref{tab:stats_mtsf}). We adopt a standard 6:2:2 train/validation/test split and report MSE and MAE as our main metrics \cite{qaio25,lee2026multivariatetimeseriesforecasting, Zhou_Zhang_Peng_Zhang_Li_Xiong_Zhang_2021}. We compare against several state-of-the-art baselines: MTSF-ANO~\cite{lee2026multivariatetimeseriesforecasting}, an ANO-based MTSF model; QuLTSF~\cite{qaio25}, a quantum method for long-term MTSF; and the strong linear baselines
DLinear/NLinear/Linear~\cite{Zeng_Chen_Zhang_Xu_2023}, which are known to
outperform many Transformer-based methods in long-term MTSF. A QFWP~\cite{10650743} baseline is also included. We set $n=C$
qubits and sweep VQC depth $D= \{1, 3\}$ and non-locality $k\in\{1,\dots,7\}$, reporting the best result over $k$ and $D$.
We train up to 100 epochs with Adam using MSE loss.

\begin{table*}[t!]
\centering
\setlength{\tabcolsep}{4pt}
\caption {\textbf{Multivariate forecasting results.} Results are averaged over 10 different random seeds. Lookback $L = 16$ and prediction horizon $H \in \{1, 5, 16, 32, 48\}$. Lower MSE and MAE are better. The best result is highlighted in \textbf{bold} and the second best is highlighted with \underline{underline}. IMP. is the improvement between the best QFWP-ANO method and the best baseline, where a larger value indicates better improvement. QFWP-ANO models implemented by us; other results from~\cite{lee2026multivariatetimeseriesforecasting}.}
\label{tab:results}
\resizebox{1.0\textwidth}{!}{
\begin{tabular}{c|c||c||wc{28pt}wc{28pt}|wc{26pt}wc{26pt}|wc{26pt}wc{26pt}||cc|cc|cc|cc|cc|cc}
\hline
\multicolumn{2}{c||}{Methods}& \multicolumn{1}{c||}{IMP.}& \multicolumn{2}{c|}{\makecell{QFWP-ANO-RO}} & \multicolumn{2}{c|}{\makecell{QFWP-ANO-R}} & \multicolumn{2}{c||}{\makecell{QFWP-ANO-O}} & \multicolumn{2}{c|}{MTSF-ANO} & \multicolumn{2}{c|}{QuLTSF} & \multicolumn{2}{c|}{QFWP} & \multicolumn{2}{c|}{DLinear} & \multicolumn{2}{c|}{NLinear} & \multicolumn{2}{c}{Linear} \\
\hline
\multicolumn{2}{c||}{Metric} & MSE & MSE & MAE & MSE & MAE & MSE & MAE & MSE & MAE & MSE & MAE & MSE & MAE & MSE & MAE & MSE & MAE & MSE & MAE \\
\hline
\multirow{5}{*}{\rotatebox{90}{ETTh1}} 
 & 1 & 3.6\% & \underline{0.133} & \underline{0.239} & 0.137 & 0.243 & \textbf{0.132} & \textbf{0.237} & 0.137 & 0.243 & 0.162 & 0.254 & 0.397 & 0.406 & 0.164 & 0.258 & 0.162 & 0.256 & 0.199 & 0.284 \\
 & 5 & 4.4\% & \underline{0.324} & \underline{0.364} & 0.337 & 0.371 & \textbf{0.323} & \textbf{0.362} & 0.338 & 0.371 & 0.431 & 0.409 & 0.700 & 0.522 & 0.478 & 0.429 & 0.489 & 0.432 & 0.510 & 0.443 \\
 & 16 & 2.3\% & \underline{0.380} & \underline{0.400} & 0.397 & 0.410 & \textbf{0.379} & \textbf{0.398} & 0.388 & 0.405 & 0.437 & 0.422 & 0.662 & 0.514 & 0.459 & 0.428 & 0.487 & 0.441 & 0.474 & 0.435 \\
 & 32 & 0.7\% & \underline{0.425} & \underline{0.426} & 0.437 & 0.432 & \textbf{0.423} & \textbf{0.425} & 0.426 & 0.427 & 0.481 & 0.447 & 0.670 & 0.523 & 0.494 & 0.447 & 0.518 & 0.458 & 0.507 & 0.453 \\
 & 48 & -2.3\% & 0.439 & 0.431 & 0.446 & 0.435 & \underline{0.437} & \underline{0.429} & \textbf{0.427} & \textbf{0.425} & 0.474 & 0.441 & 0.645 & 0.514 & 0.481 & 0.439 & 0.502 & 0.450 & 0.490 & 0.443 \\
\hline
\multirow{5}{*}{\rotatebox{90}{ETTh2}} 
 & 1 & -1.4\% & \underline{0.075} & 0.170 & 0.082 & 0.179 & \underline{0.075} & 0.169 & 0.079 & 0.173 & \textbf{0.074} & \underline{0.167} & 0.117 & 0.225 & \textbf{0.074} & \textbf{0.166} & \textbf{0.074} & \textbf{0.166} & 0.076 & 0.170 \\
 & 5 & 2.5\% & \textbf{0.118} & \textbf{0.215} & 0.129 & 0.229 & \textbf{0.118} & \textbf{0.215} & \underline{0.121} & \underline{0.218} & 0.122 & 0.222 & 0.155 & 0.261 & 0.126 & 0.225 & 0.126 & 0.226 & 0.127 & 0.228 \\
 & 16 & 2.3\% & \underline{0.174} & \underline{0.262} & 0.177 & 0.265 & \textbf{0.173} & \textbf{0.261} & 0.180 & 0.268 & 0.177 & 0.271 & 0.193 & 0.284 & 0.177 & 0.269 & 0.179 & 0.271 & 0.178 & 0.270 \\
 & 32 & 0.5\% & \underline{0.222} & \underline{0.293} & 0.223 & 0.294 & \textbf{0.221} & \textbf{0.292} & 0.225 & 0.295 & \underline{0.222} & 0.297 & 0.243 & 0.314 & \underline{0.222} & 0.294 & 0.223 & 0.296 & \underline{0.222} & 0.295 \\
 & 48 & 0.0\% & \underline{0.261} & \underline{0.315} & \underline{0.261} & 0.316 & \textbf{0.259} & \textbf{0.314} & 0.264 & 0.318 & \textbf{0.259} & 0.318 & 0.277 & 0.331 & \textbf{0.259} & \underline{0.315} & \underline{0.261} & 0.317 & \textbf{0.259} & 0.316 \\
\hline
\multirow{5}{*}{\rotatebox{90}{ETTm1}} 
 & 1 & 0.0\% & \textbf{0.048} & \underline{0.136} & \underline{0.050} & 0.138 & \textbf{0.048} & \underline{0.136} & \textbf{0.048} & \textbf{0.135} & 0.051 & \underline{0.136} & 0.105 & 0.201 & 0.052 & 0.138 & 0.052 & 0.137 & 0.052 & 0.138 \\
 & 5 & 3.6\% & \textbf{0.106} & \textbf{0.200} & 0.111 & 0.205 & \underline{0.107} & \textbf{0.200} & 0.110 & \underline{0.204} & 0.128 & 0.210 & 0.198 & 0.271 & 0.133 & 0.214 & 0.132 & 0.214 & 0.133 & 0.214 \\
 & 16 & 5.2\% & \textbf{0.311} & \underline{0.332} & \underline{0.324} & 0.341 & \textbf{0.311} & \textbf{0.331} & 0.328 & 0.342 & 0.432 & 0.369 & 0.520 & 0.423 & 0.451 & 0.378 & 0.451 & 0.378 & 0.452 & 0.378 \\
 & 32 & 5.7\% & \textbf{0.560} & \textbf{0.448} & 0.575 & 0.459 & \underline{0.564} & \underline{0.451} & 0.594 & 0.462 & 0.788 & 0.517 & 0.859 & 0.552 & 0.835 & 0.534 & 0.836 & 0.534 & 0.836 & 0.534 \\
 & 48 & 6.6\% & \textbf{0.660} & \textbf{0.501} & \underline{0.668} & 0.507 & \underline{0.668} & \underline{0.505} & 0.720 & 0.522 & 0.956 & 0.588 & 0.986 & 0.608 & 1.018 & 0.610 & 1.019 & 0.610 & 1.019 & 0.610 \\
\hline
\multirow{5}{*}{\rotatebox{90}{ETTm2}} 
 & 1 & -3.1\% & \underline{0.033} & 0.103 & 0.035 & 0.107 & \underline{0.033} & 0.103 & 0.034 & 0.106 & \textbf{0.032} & \textbf{0.095} & 0.051 & 0.136 & \textbf{0.032} & \textbf{0.095} & \textbf{0.032} & \underline{0.096} & \textbf{0.032} & \underline{0.096} \\
 & 5 & -1.7\% & \underline{0.059} & \underline{0.140} & 0.060 & 0.143 & \underline{0.059} & \underline{0.140} & \underline{0.059} & 0.141 & \textbf{0.058} & \textbf{0.138} & 0.073 & 0.166 & 0.060 & 0.142 & 0.060 & 0.142 & 0.060 & 0.142 \\
 & 16 & 1.0\% & \textbf{0.099} & \textbf{0.190} & 0.101 & 0.193 & \textbf{0.099} & \textbf{0.190} & \underline{0.100} & \underline{0.191} & 0.104 & 0.196 & 0.118 & 0.216 & 0.108 & 0.201 & 0.108 & 0.201 & 0.108 & 0.201 \\
 & 32 & 0.7\% & \underline{0.146} & \underline{0.237} & 0.149 & 0.241 & \textbf{0.145} & \textbf{0.236} & \underline{0.146} & 0.238 & 0.158 & 0.250 & 0.168 & 0.263 & 0.162 & 0.255 & 0.162 & 0.255 & 0.162 & 0.255 \\
 & 48 & 0.6\% & \underline{0.179} & \underline{0.266} & 0.183 & 0.271 & \textbf{0.178} & \textbf{0.265} & \underline{0.179} & 0.267 & 0.195 & 0.284 & 0.205 & 0.295 & 0.198 & 0.288 & 0.198 & 0.288 & 0.198 & 0.288 \\
\hline
\end{tabular}
}
\end{table*}

\vspace{-0.5em}
\subsection{Reinforcement Learning}

Following Lin et al.~\cite{11250031}, we embed QFWP-ANO as the function approximator in Asynchronous Advantage Actor-Critic (A3C)~\cite{pmlr-v48-mniha16}: a linear layer encodes the state $s_t$ into a latent $\boldsymbol{h}_t$, which is the circuit input for two independent QFWP-ANO instances (policy and value) that produce $\pi_\theta(a\mid s_t)$ and $V_\psi(s_t)$. Each QFWP-ANO instance's hypernetwork reads $s_t$ directly to generate the parameters.

Following prior work~\cite{NEURIPS2021_eec96a7f, Chen:2020opi, 11250031}, we evaluate on Acrobot, a swing-up control problem, and MiniGrid-SimpleCrossingS9N1, a sparse-reward navigation task (Table~\ref{tab:stats_rl}). We compare our method against an ANO-VQC baseline. In all configurations, we use $n=4$ qubits, $D=4$ VQC layers, and non-locality $k=3$. Further, we introduce a trainable input scaling $\boldsymbol{w}$ (e.g., $R_y(w_i x_i)$ rather than $R_y(x_i)$). Since trainable input scaling $\boldsymbol{w}$ has proven beneficial in standard QRL agents~\cite{NEURIPS2021_eec96a7f,lee2026quantumhierarchicalreinforcementlearning}, we evaluate models both with and without $\boldsymbol{w}$ to investigate how it affects ANO-based agents. All models are trained using A3C for 5{,}000 (Acrobot) or 10{,}000 (SimpleCrossingS9N1) episodes, averaged over 10 random seeds.

\section{Experiments}

\begin{table}[t!]
\centering
\setlength{\tabcolsep}{4pt}
\caption{\textbf{Impact of ANO non-locality and VQC depth.} Color shows the best variant among {\color{ForestGreen}R}/{\color{red}O}/{\color{Plum}RO}. \textbf{Bold} indicates lowest MSE across all $k=1$ to $k = 7$ within that column, i.e., the best variant and non-locality $k$ for a given ($H$, $D$) pair.}
\label{tab:ablation_locality}
\begin{subtable}{\columnwidth}
\centering
\caption{VQC depth $D=1$.}
\begin{tabular}{cccccc}
\toprule
Settings & $H = 1$ & $H = 5$ & $H = 16$ & $H = 32$ & $H = 48$ \\
\midrule
$k=1$ & \textcolor{red}{0.073} & \textcolor{red}{0.161} & \textcolor{red}{0.260} & \textcolor{red}{0.359} & \textcolor{red}{0.404} \\
$k=2$ & \textcolor{red}{\textbf{0.072}} & \textcolor{red}{\textbf{0.152}} & \textcolor{red}{0.242} & \textcolor{Plum}{0.340} & \textcolor{red}{0.388} \\
$k=3$ & \textcolor{red}{0.073} & \textcolor{red}{\textbf{0.152}} & \textcolor{red}{\textbf{0.240}} & \textcolor{Plum}{\textbf{0.338}} & \textcolor{Plum}{\textbf{0.385}} \\
$k=4$ & \textcolor{red}{0.074} & \textcolor{red}{\textbf{0.152}} & \textcolor{red}{0.242} & \textcolor{Plum}{0.340} & \textcolor{Plum}{0.388} \\
$k=5$ & \textcolor{red}{0.074} & \textcolor{red}{0.155} & \textcolor{red}{0.248} & \textcolor{ForestGreen}{0.347} & \textcolor{ForestGreen}{0.391} \\
$k=6$ & \textcolor{red}{0.077} & \textcolor{ForestGreen}{0.168} & \textcolor{ForestGreen}{0.263} & \textcolor{ForestGreen}{0.360} & \textcolor{ForestGreen}{0.404} \\
$k=7$ & \textcolor{ForestGreen}{0.179} & \textcolor{ForestGreen}{0.268} & \textcolor{ForestGreen}{0.379} & \textcolor{ForestGreen}{0.475} & \textcolor{ForestGreen}{0.519} \\
\bottomrule
\end{tabular}
\end{subtable}

\vspace{1em}

\begin{subtable}{\columnwidth}
\centering
\caption{VQC depth $D=3$.}
\begin{tabular}{cccccc}
\toprule
Settings & $H = 1$ & $H = 5$ & $H = 16$ & $H = 32$ & $H = 48$ \\
\midrule
$k=1$ & \textcolor{red}{0.073} & \textcolor{red}{0.160} & \textcolor{red}{0.260} & \textcolor{red}{0.359} & \textcolor{red}{0.403} \\
$k=2$ & \textcolor{red}{\textbf{0.072}} & \textcolor{red}{\textbf{0.152}} & \textcolor{red}{0.243} & \textcolor{red}{0.341} & \textcolor{red}{0.388} \\
$k=3$ & \textcolor{red}{0.073} & \textcolor{red}{0.153} & \textcolor{red}{\textbf{0.242}} & \textcolor{red}{\textbf{0.340}} & \textcolor{red}{\textbf{0.386}} \\
$k=4$ & \textcolor{red}{0.074} & \textcolor{red}{0.153} & \textcolor{red}{0.243} & \textcolor{red}{0.342} & \textcolor{red}{0.388} \\
$k=5$ & \textcolor{red}{0.076} & \textcolor{red}{0.155} & \textcolor{red}{0.251} & \textcolor{red}{0.350} & \textcolor{ForestGreen}{0.395} \\
$k=6$ & \textcolor{red}{0.079} & \textcolor{ForestGreen}{0.168} & \textcolor{ForestGreen}{0.267} & \textcolor{ForestGreen}{0.363} & \textcolor{ForestGreen}{0.408} \\
$k=7$ & \textcolor{ForestGreen}{0.167} & \textcolor{ForestGreen}{0.277} & \textcolor{ForestGreen}{0.374} & \textcolor{ForestGreen}{0.475} & \textcolor{ForestGreen}{0.520} \\
\bottomrule
\end{tabular}
\end{subtable}
\end{table}

\subsection{MTSF Results}
Table~\ref{tab:results} summarizes the MTSF performance at a fixed lookback $L=16$. QFWP-ANO-O achieves the lowest MSE in 13 of the 20 settings; combined with QFWP-ANO-RO, they rank first in 16 settings and secure the second-lowest MSE in the remaining four. This advantage is largest and most consistent on ETTm1, where the improvement over the best baseline by up to 6.6\% and the improvement grows steadily with the horizon. Further, QFWP-ANO outperforms standard ANO-VQCs (MTSF-ANO) in 17 of 20 cases. Across all ETT datasets, programming the observable (QFWP-ANO-O, QFWP-ANO-RO) is more effective than programming only the rotation angles (QFWP-ANO-R). Our finding indicate that input-conditioned non-local measurement provides significant gains.

\vspace{-0.5em}
\subsection{RL Results}

\begin{figure}[t!]
\centering
\includegraphics[width=\columnwidth]{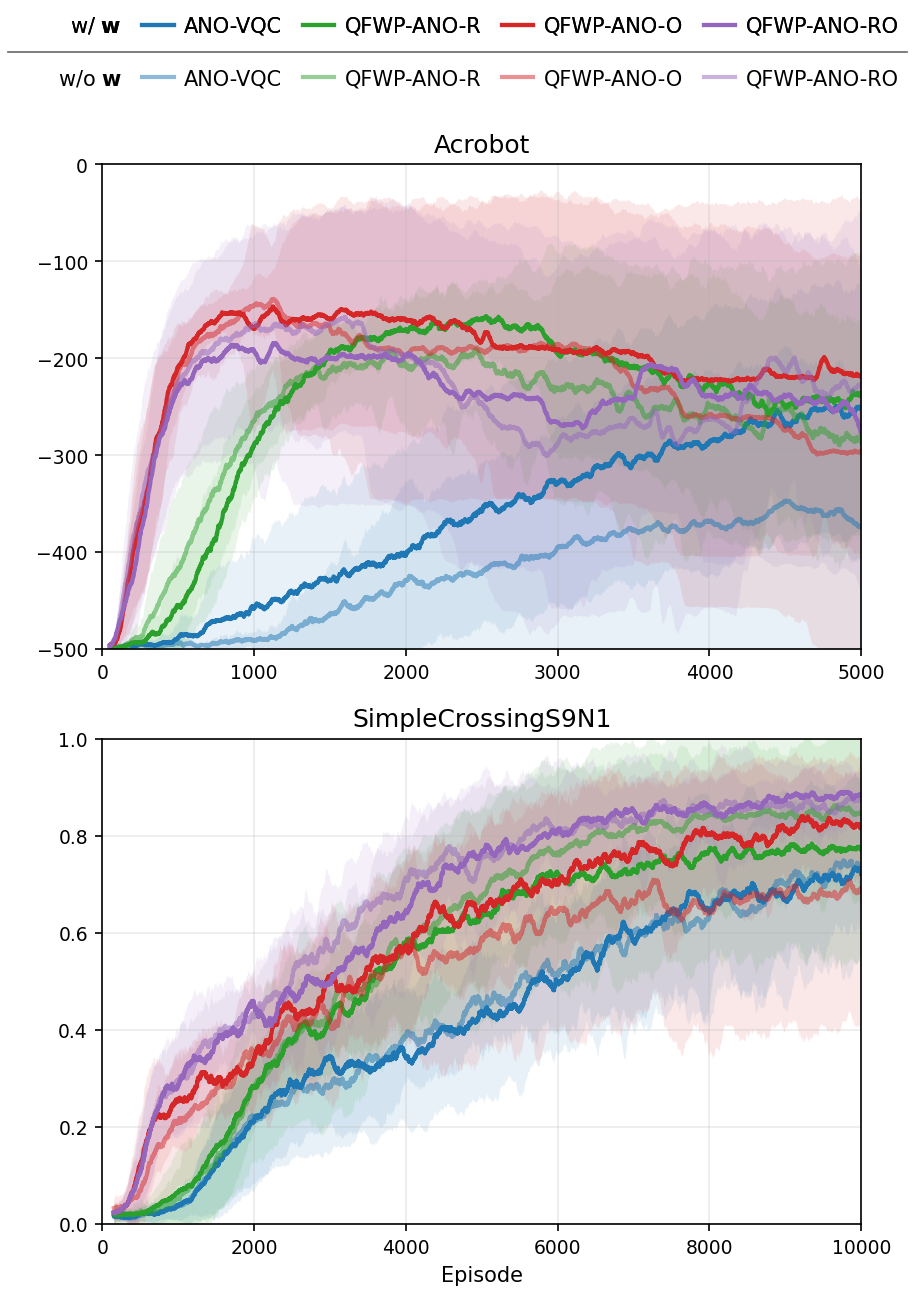}
\caption{\textbf{Reward over episodes.} Band is $\pm$ standard deviation.}
\label{fig:rl_curves}
\end{figure}

Fig.~\ref{fig:rl_curves} shows the learning curves, with (dark) and without (light) the trainable input scaling $\boldsymbol{w}$. In Acrobot, all three QFWP-ANO variants substantially outperform the ANO-VQC baseline from episode 1000 to 4000. Although the models ultimately converge to similar final returns, QFWP-ANO-O and QFWP-ANO-RO reach their peak performance as early as around episode 1000. This early convergence demonstrates that programming the observable effectively improves sample efficiency. In SimpleCrossingS9N1, QFWP-ANO-RO reaches the highest final reward and remains the best throughout training, ahead of all other models. Employing the trainable input scaling $\boldsymbol{w}$ only significantly boosts ANO-VQC in Acrobot.

\vspace{-0.5em}
\subsection{Impact of ANO Non-Locality and VQC Depth}

We evaluate how VQC depth and ANO non-locality influence MTSF performance of QFWP-ANO variants. Table~\ref{tab:ablation_locality} reports the lowest average MSE across the four ETT datasets for each $(k,H)$ configuration and circuit depth $D \in \{1,3\}$, colored by the best-performing QFWP-ANO variant. We make four key observations: (1) MSE initially decreases from $k=1$, reaches a minimum around $k \in \{2, 3\}$, and rises sharply by $k=7$, which is likely due to combinatorial ANO scheme leaving only a single expectation value at $k = 7$; (2) QFWP-ANO-RO and QFWP-ANO-O achieve the lowest MSE for $k \leq 4$, while QFWP-ANO-R starting to overtake them from $k=5$ and dominates across all horizons at $k=7$; (3) moderate non-locality consistently yields the best performance, regardless of the specific variant; and (4) shallower circuits ($D=1$) match or outperform deeper ones ($D=3$) in most settings.

\section{Conclusion}
We introduced QFWP-ANO, a novel QNN architecture
that utilizes a classical hypernetwork to dynamically
program VQCs along with non-local observables. Across several MTSF and RL benchmarks, programming the non-local observables yields the most substantial improvements. Specifically, in MTSF, QFWP-ANO ranks first in MSE in 16 of 20 settings and ranks second in the remaining four. In RL environments, QFWP-ANO consistently outperforms ANO-VQCs, and programming the non-local observables can effectively accelerate learning. These findings establish input-conditioned ANO as a promising route to more capable quantum models.



\vfill\pagebreak

\ninept
\bibliographystyle{IEEEbib}
\bibliography{reference}

\end{document}